\documentclass{article}

\usepackage{microtype}
\usepackage{graphicx}
\usepackage{booktabs} 

\usepackage{hyperref}

\usepackage[accepted]{icml2020}

\icmltitlerunning{Incidence Networks for Geometric Deep Learning}

\usepackage[utf8]{inputenc} 
\usepackage[T1]{fontenc}    
\hypersetup{
    colorlinks = false,
    linkcolor = red,
    citecolor = red,
    citebordercolor = {0 1 0},
    linkbordercolor = {1 0 0},
}
\usepackage{url}            
\usepackage{amsfonts}       
\usepackage{nicefrac}       
\usepackage{xcolor}
\usepackage{amsmath,amsthm,amssymb}
\usepackage{mathtools}
\usepackage{tikz}
\usetikzlibrary{arrows,automata,positioning,arrows.meta,bending,matrix,shapes}
\usepackage{MnSymbol}
\usepackage{wasysym}
\usepackage{multirow} 
\usepackage{xspace}
\usepackage{dsfont}
\usepackage[cal=dutchcal,
calscaled=1,
bb=boondox,
scr=euler
]{mathalfa}
\usepackage{wrapfig}
\usepackage{verbatim}
\usepackage[capitalise]{cleveref}
\crefformat{equation}{(#2#1#3)}
\usepackage{sidecap}
\usepackage{subcaption}
\usepackage{footnote}
\usepackage[framemethod=TikZ]{mdframed}
\usepackage{pifont}
\usepackage{abraces}

\usepackage{bm}

\newtheorem{theorem}{Theorem}[section]

\newtheorem{definition}{Definition}

\newtheorem{example}{Example}

\mdfdefinestyle{MyFrame}{%
    linecolor=white,
    outerlinewidth=1pt,
    roundcorner=2pt,
    innertopmargin=\baselineskip,
    innerbottommargin=\baselineskip,
    innerrightmargin=5pt,
    innerleftmargin=5pt,
    backgroundcolor=black!3!white}

\newcommand{\ie}[0]{\emph{i.e.},~}
\newcommand{\eg}[0]{\emph{e.g.},~}

\newcommand{\Pool}{\operatorname{Pool}}
\newcommand{\Bcast}{\operatorname{Bcast}}
\newcommand{\Aut}{\gr{Aut}}

\newcommand{\mat}[1]{\ensuremath{{\mathbf{#1}}}}
\newcommand{\tens}[1]{\ensuremath{{\mathbf{#1}}}}

\newcommand{\set}[1]{\ensuremath{\mathbb{#1}}}
\newcommand{\gr}[1]{\ensuremath{\mathcal{#1}}}
\newcommand{\tuple}[1]{\ensuremath{\langle{#1} \rangle}}
\newcommand{\Real}{\mathbb{R}}
\newcommand{\Nat}{\mathbb{N}}

\newcommand{\W}{\mathsf{W}}
\newcommand{\w}{{\mathsf{w}}}
\newcommand{\B}{\mat{Y}}
\newcommand{\A}{\mat{X}}

\newcommand{\X}{\tens{X}}

\newcommand{\Y}{\tens{Y}}

\newcommand{\nf}{n}
\newcommand{\ef}{e}
\newcommand{\nfb}{\boldsymbol{n}}
\newcommand{\efb}{\boldsymbol{e}}

\newcommand{\deltas}{\boldsymbol{\delta}}
\newcommand{\sigmas}{\boldsymbol{\sigma}}
   
\newcommand{\st}{\;|\;}

\newcommand{\stirling}{\genfrac{\{}{\}}{0pt}{}}

\newcommand{\myset}[1]{\ensuremath{\{#1}\}}
\newcommand{\Ga}{\W}
\newcommand{\Gb}[2]{\W^{#1\to #2}}
\newcommand{\Gc}[3]{\W^{#1\to #2}_{#3}}
\newcommand{\sm}[1]{\scriptstyle{\ensuremath{#1}}}
\newif\ifstartedinmathmode
\newcommand\encircled[1]{%
  \relax\ifmmode\startedinmathmodetrue\else\startedinmathmodefalse\fi%
  \tikz[baseline,anchor=base]{%
  \node[draw,circle,outer sep=0pt,inner sep=.2ex]
    {\ifstartedinmathmode$#1$\else#1\fi};}%
}
\newif\ifstartedinmathmode
\newcommand\ellipse[1]{%
  \relax\ifmmode\startedinmathmodetrue\else\startedinmathmodefalse\fi%
  \tikz[baseline,anchor=base]{%
  \node[draw,ellipse,outer sep=0pt,inner sep=0pt, minimum height=12pt]
    {\ifstartedinmathmode$#1$\else#1\fi};}%
}

\makesavenoteenv{tabular}
\makesavenoteenv{table}

\newbool{APX}
\booltrue{APX}

\begin{document}
\twocolumn[
\icmltitle{Incidence Networks for Geometric Deep Learning}



\icmlsetsymbol{equal}{*}

\begin{icmlauthorlist}
\icmlauthor{Marjan Albooyeh}{equal,ubc}
\icmlauthor{Daniele Bertolini}{equal}
\icmlauthor{Siamak Ravanbakhsh}{mcgill,mila}
\end{icmlauthorlist}

\icmlaffiliation{ubc}{Department of Computer Science, University of British Columbia, Vancouver, Canada}
\icmlaffiliation{mcgill}{School of Computer Science, McGill University, Montreal, Canada}
\icmlaffiliation{mila}{Mila Quebec AI Institute, Montreal, Canada}

\icmlcorrespondingauthor{Marjan Albooyeh}{albooyeh@cs.ubc.ca}
\icmlcorrespondingauthor{Daniele Bertolini}{dbertolini84@gmail.com}

\icmlkeywords{Machine Learning, ICML}

\vskip 0.3in
]



\printAffiliationsAndNotice{\icmlEqualContribution} 

\begin{abstract}
Sparse incidence tensors can represent a variety of structured data. For example, we may represent attributed graphs using their node-node, node-edge, or edge-edge 
incidence matrices. In higher dimensions, incidence tensors can represent simplicial complexes and polytopes.
In this paper, we formalize incidence tensors, analyze their structure, and present the family of equivariant networks that operate on them.
We  show  that  any  incidence tensor decomposes into invariant subsets. 
This decomposition, in turn, leads to a decomposition of the corresponding equivariant linear maps. We characterize these linear maps as a combination of efficient pooling-and-broadcasting operations.
\end{abstract}

\begin{figure*}[ht]
\centering
\includegraphics[width=\linewidth]{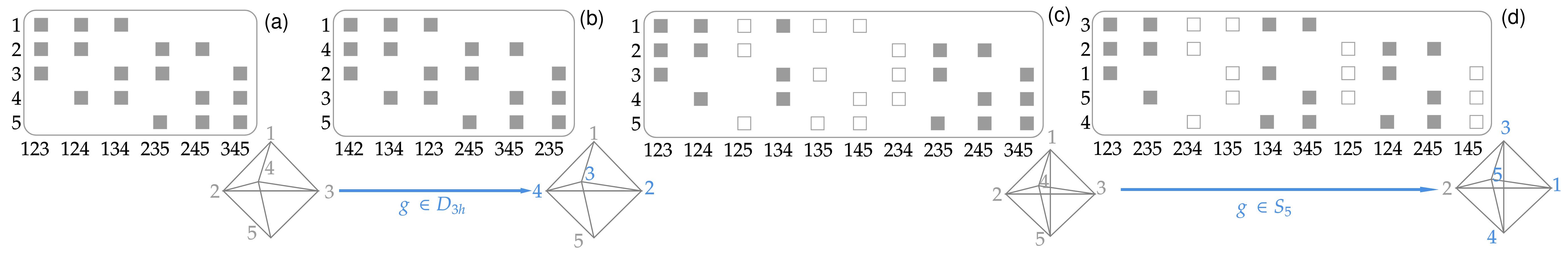}
\caption{\footnotesize{
\textbf{a}) The sparsity pattern in the node-face incidence matrix for an (undirected) \emph{triangular bi-pyramid} (concatenation of two tetrahedra). Note that each face (column) is adjacent to exactly three nodes.
\textbf{b}) Nodes are permuted using a member of the symmetry group of the object $\pi \in \gr{D}_{3h} \leq \gr{S}_5$. This permutation of nodes imposes a natural permutation action on the faces in which $\{\delta_1, \delta_2, \delta_3\} \mapsto \{ \pi \cdot \delta_1, \pi \cdot \delta_2, \pi \cdot \delta_3\}$. Note that permutations from the automorphism group preserve the sparsity pattern of the incidence matrix.
\textbf{c}) The geometric object of (a) after \textit{densification}: the incidence matrix now includes all possible faces of size three,
however, it still maintains a specific sparsity pattern.
\textbf{d}) After densifying the structure, \textit{any} permutation of nodes (and corresponding permutation action on faces of the dense incidence matrix) preserves its sparsity pattern.
}}
\label{fig:tetra}
\end{figure*}

\section{Introduction}
\label{sec:intro}
Many interesting data structures can be represented with sparse incidence tensors.
For example, we can represent graphs using both node-node and node-edge sparse incidence matrices. 
We can extend this incidence representation to data defined on
simplicial complexes and polytopes of arbitrary dimension, such as mesh, polygons, and polyhedra. 
The goal of this paper is to design deep models for these structures.

We represent an attributed geometric structure using its \emph{incidence tensor}, which models the incidence {pattern} of its \emph{faces}.
For example, rows and columns in a node-edge incidence matrix are indexed by faces of size one  (nodes) and two (edges). 
Moreover each edge (column) is incident to exactly two nodes (rows).
The sparsity pattern of the incidence tensor has important information about the geometric structure. This is because sparsity preserving permutation of nodes often match the \emph{automorphism group} of the geometric object; see \cref{fig:tetra}(a,b).

We are interested in designing models that are informed by the symmetry of the underlying structure. We do so by making the model equivariant to symmetry transformations. 
When using the incidence tensor representation, a natural choice of symmetry transformations is the {automorphism group} of the geometric object. 
However, when working with a dataset comprising of different instances (\eg different graphs or polyhedra), using individual automorphism groups is not practical. This is because each symmetry group dictates a different equivariant model, and we cannot train a single model on the whole dataset.
A solution is to use  the \emph{symmetric group} (the group of all permutations of nodes) 
for all instances, which implicitly assumes a dense structure where all faces are present, \eg all graphs are fully connected; see \cref{fig:tetra}(c,d). 

We show that under the action of the symmetric group, any incidence tensor decomposes into {invariant subsets}, or \emph{orbits}, where each orbit corresponds to 
faces of particular size.
For example, a node-node incidence matrix decomposes into: 1) diagonals, that can encode node attributes, and; 2) off-diagonals, corresponding to edge attributes. This is because any  permutation of nodes, (\ie simultaneous permutation of rows and columns) moves an (off-) diagonal entry to another (off-) diagonal entry in the node-node incidence.
We can vectorize the diagonal and off-diagonal entries to get a \emph{node vector}
and an \emph{edge vector}. These are examples of \emph{face-vectors} in the general setting, and this example shows how and incidence tensor decomposes into face-vectors.

This decomposition into face-vectors also breaks up the design of equivariant linear maps for arbitrary incidence tensors into design of such maps between face-vectors of different size.
We show that any such linear map can be written as a linear combination of efficient pooling-and-broadcasting operations.
These equivariant linear maps replace the linear layer in a feedforward neural network to create an \emph{incidence network}.

\section{Related Works}\label{sec:related}
Deep learning with structured data is a very active area of research.
Here, we briefly review some of the closely related works in graph learning and equivariant deep learning.

\paragraph{\textsc{Graph Learning.}} 
The idea of graph neural networks goes back to the work of \cite{scarselli2009graph}.
More recently, \citet{gilmer2017neural} introduced the message passing neural networks and showed that they subsume several other graph neural network architectures~\cite{li2015gated,duvenaud2015convolutional,kearnes2016molecular,schutt2017quantum}, including the spectral methods that follows.
Another body of work in graph deep learning extends convolution to graphs using the spectrum of the graph Laplacian~\cite{bronstein2017geometric,bruna2013spectral}. While principled, in its complete form, the Fourier bases extracted from the Laplacian are instance-dependent and the lack of any parameter or function sharing across the graphs limits their generalization. 
Following \cite{henaff2015deep, defferrard2016convolutional}, \citet{kipf2016semi} propose a single-parameter simplification of spectral method that addresses this limitation and it is widely used in practice. Some notable extensions and related ideas include \cite{velivckovic2017graph,hamilton2017inductive,xu2018representation,zhang2018end,ying2018hierarchical,morris2018weisfeiler,maron2019provably}.
 
\paragraph{\textsc{Equivariant Deep Learning.}} 
Equivariance constrains the predictions of a model $\phi: \set{X} \mapsto \set{Y}$ under a group $\gr{G}$ of transformations of the input, such that
\begin{equation}
\label{eq:equiv}
\phi(\pi \cdot x) = \pi \cdot \phi(x),\, \forall x \in \set{X}, \forall\pi \in \gr{G}.
\end{equation}
Here $\pi \cdot x$ is a consistently defined transformation of $x$ parameterized by $\pi \in \gr{G}$, while $\pi \cdot \phi(x)$ denotes the corresponding transformation of the output.
For example, in a convolution layer \cite{lecun1998gradient}, $\gr{G}$ is the group of discrete translations, and \cref{eq:equiv} means that any translation of the input leads to the same translation of the output. When $\phi: x \mapsto \sigma(\W x)$ is a standard feed-forward layer with parameter matrix $\W$, the equivariance property \cref{eq:equiv} enforces parameter-sharing in $\W$ \cite{shawe1993symmetries,ravanbakhsh_symmetry}.

Most relevant to our work, are equivariant models proposed for geometric deep learning that we review next.
Covariant compositional network \cite{kondor2018covariant} extends the message passing framework by considering basic tensor operations that preserve equivariance.
While the resulting architecture can be quite general, it comes at the cost of efficiency. 
\citet{hartford2018deep} propose a linear map equivariant to independent permutation of different dimensions of a tensor. 
\emph{Equivariant graph networks} of \cite{maron2018invariant} model the interactions within a set of nodes. We will further discuss this model as a special type of incidence network.
These equivariant layers for interactions between and within sets are further generalized to multiple types of interactions in \cite{graham2019deep}.
Several recent works investigate the universality of such equivariant networks \cite{maron2019universality,keriven2019universal,chen2019equivalence}.
A flexible approach to equivariant and geometric deep learning where a global symmetry is lacking is proposed in \cite{cohen2019gauge}.

\begin{figure*}[ht]
\centering
\includegraphics[width=\linewidth]{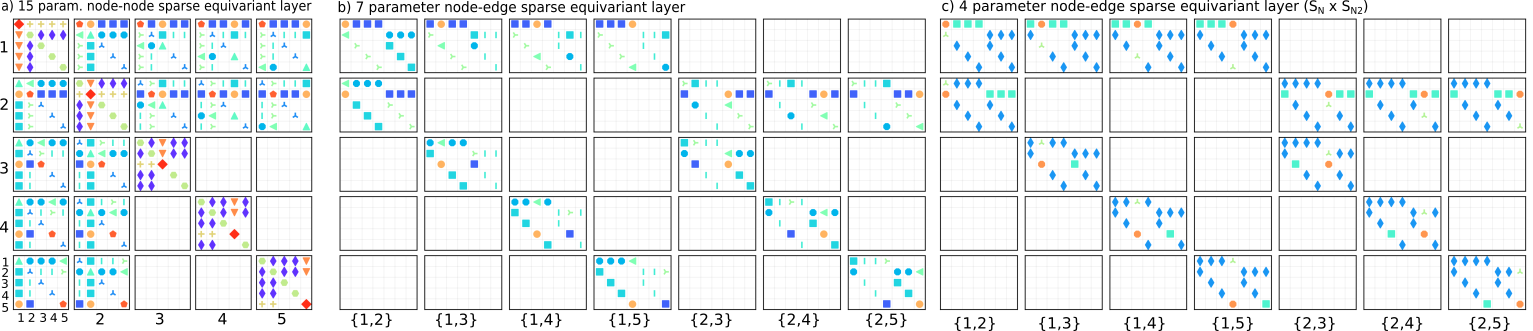}
\caption{\footnotesize{
Parameter-sharing in the receptive field of  equivariant map: (\textbf{left}) 15 parameter layer for node-node incidence, (\textbf{middle}) 7 parameter layer for node-edge incidence, and (\textbf{right}) 4 parameter layer node-edge incidence  \textbf{\textsc{Block structure.}} The adjacency structure of the undirected graph with 5 nodes and 7 edges is evident from the sparsity patterns. Here, each inner block shows the parameter-sharing in the receptive field of the corresponding output unit. For example, the block on row 1 and column $\{1,3\}$ of the (middle) figure shows the dependency of the output incidence matrix at that location on the entire input incidence matrix. \textbf{\textsc{Decomposition.}} The total number of unique parameters in (left) is 15 compared to 7 for the (middle). As shown in \cref{sec:graph-linear-layers} the 15 ($=7 + 2 + 3 + 3$) parameter model decomposes into 4 linear maps, one of which is isomorphic to the 7 parameter model. One could also identify the 7 unique symbols of (middle) in the parameter-sharing of the (left). Note that these symbols appear on the off-diagonal blocks and off-diagonal elements within blocks, corresponding to input and output \emph{edges}.
}
}
\label{fig:layer}
\end{figure*}

\section{Graphs}
\label{sec:graphs}
In this section we discuss graphs and later generalize the arguments to a broader set of geometric objects in \cref{sec:higher-order-objects}. Without loss of generality, we assume a fully-connected graph $G = ([N], \set{E}\subseteq [N] \times [N])$, where $[N] = \{1,\ldots,N\}$ denotes a set of nodes, and $\set{E}$ a set of edges.

\subsection{Linear Layers}
\label{sec:graph-linear-layers}
There are two common ways of representing $G$. The first approach is to use a node-node incidence matrix $\X_{\delta_1,\delta_2}$ indexed by $\delta_1,\delta_2 \in [N]$. In this representation, node and edge features are encoded as diagonal and off-diagonal entries of $\X$, respectively. Here, we assume single input and output channel (\ie scalar node and edge attributes) for simplicity; results trivially generalize to multiple channels.

Consider the group of all $N!$ permutations of nodes $\gr{S}_{N}$ and its \emph{action} on $\X$, which simultaneously permutes the rows and columns of $\X$. Let $\W: \Real^{N \times N} \to \Real^{N \times N}$ be a linear map equivariant to this action of $\gr{S}_N$,
\begin{equation}
\label{eq:equivariance-X}
\W(\pi\X\pi^\top) = \pi\W(\X)\pi^\top,\quad\forall \pi\in\gr{S}_{N}, \forall\X.
\end{equation}
The map $\W$ is constrained so that permuting the rows and columns of the input will have the same effect on the output.
As shown in \cite{maron2018invariant} this condition constrains the number of independent parameters in $\W$ to fifteen, regardless of the size of the graph. 

Alternatively, one can represent $G$ with a node-edge incidence matrix $\Y_{\delta_1,\{\delta_2,\delta_3\}}$ where $\delta_1 \in [N]$ labels nodes and the unordered pair $\{\delta_2,\delta_3\}$ with $\delta_2 \neq \delta_3 \in [N]$ labels edges.
$\Y$ has a special sparsity pattern: $\Y_{\delta_1,\{\delta_2,\delta_3\}} \neq 0$ iff node $\delta_1$ is incident to the edge $\{\delta_2,\delta_3\}$. We identify this sparsity pattern implicitly by writing the matrix as $\Y_{\delta_1,\{\delta_1,\delta_2\}}$, so that we only index non-zero entries (note the repeated $\delta_1$).
$\Y$ naturally encodes edge features  at $\Y_{\delta_1,\{\delta_1,\delta_2\}}$ and $\Y_{\delta_2,\{\delta_1,\delta_2\}}$ for two different edge-directions of the edge $\{\delta_1,\delta_2\}$. 

The action of $\gr{S}_{N}$ on $\Y$ is also a simultaneous permutation of rows and columns, where the permutation of columns is defined by the action on the node pair that identifies each edge. 
$\pi \cdot \{\delta_1,\delta_2\} = \{\pi \cdot \delta_1, \pi \cdot \delta_2\}$.
This action preserves the sparsity pattern of $\Y$
defined above (note that even a fully connected graph has a sparse node-edge incidence matrix.) The maximal equivariant linear map acting on $\Y$ is constrained to have seven independent parameters (assuming a single input and output channel). 
More alternatives beside node-node and node-edge representation for graph exist -- \eg one may use an edge-edge incidence matrix. 


Since both $\X$ and $\Y$ represent the same graph $G$, and the corresponding linear maps are equivariant to the same group $\gr{S}_{N}$, one expects a relationship between the two representations and maps. This relationship is due to decomposition of $\X$ into orbits under the action of $\gr{S}_{N}$. In particular, $\X$ decomposes into two orbits: diagonal elements ($\X_{\{\delta_1\}}$) and  off-diagonal elements ($\X_{\{\delta_1,\delta_2\}}$), where each subset is invariant under the action of $\gr{S}_{N}$ -- that is simultaneous permutations of rows and columns do not move a diagonal element to off-diagonal or vice-versa. We write this decomposition as 
\begin{equation}
\X \cong \X_{\{\delta_1\}} \cupdot \X_{\{\delta_1,\delta_2\}},
\end{equation}
where the diagonal orbit is isomorphic to the vector of nodes $\X_{\{\delta_1\}}$ and the off-diagonal orbit is isomorphic to the vector of edges $\X_{\{\delta_1,\delta_2\}}$ with $\delta_1 \neq \delta_2$. Consider the map $\W$ defined above, in which both input and target decompose in this way. It follows that the map itself also decomposes into four maps
\begin{equation}
\label{eq:graph-map-decomposition}
\W(\X) = \bigcupdot_{m'=1}^{2}\left(\W^{1\to m'}(\X_{\{\delta_1\}})+\W^{2\to m'}(\X_{\{\delta_1,\delta_2\}})\right),
\end{equation}
where $\W^{m \to m'}$ maps a face-vector of faces of size $m$ to a face-vector of faces of size $m'$.
Equivariance to $\gr{S}_N$ for each of these maps constrains the number of independent parameters: $\W^{1\to 1}$ is the equivariant layer used in DeepSets \cite{zaheer_deepsets}, and has two parameters. $\W^{1\to 2}$ and $\W^{2\to 1}$ each have three parameters, and $\W^{2\to 2}$ has seven unique parameter, and maps input edge features into target edge features. One key point is that the edge-vector $\X_{\{\delta_1,\delta_2\}}$ is isomorphic to the node-edge incidence matrix $\Y_{\delta_1,\{\delta_1,\delta_2\}}$ and thus the seven-parameters equivariant map for $\Y$ is exactly $\W^{2\to 2}(\X_{\{\delta_1,\delta_2\}})$ of \cref{eq:graph-map-decomposition}. The parameter-sharing in two linear maps for the node-node and node-edge incidence matrices are visualized in \cref{fig:layer}\,(left, middle).
\footnote{As a side note, one can also encode node features in a node-edge incidence matrix by doubling the number of channels and broadcasting node-features across all edges incident to a node. In this case all fifteen operations are retrieved, and the two layers for $\X$ and $\Y$ are equivalent. }

\begin{figure}[ht]
\centering
\includegraphics[width=.9\linewidth]{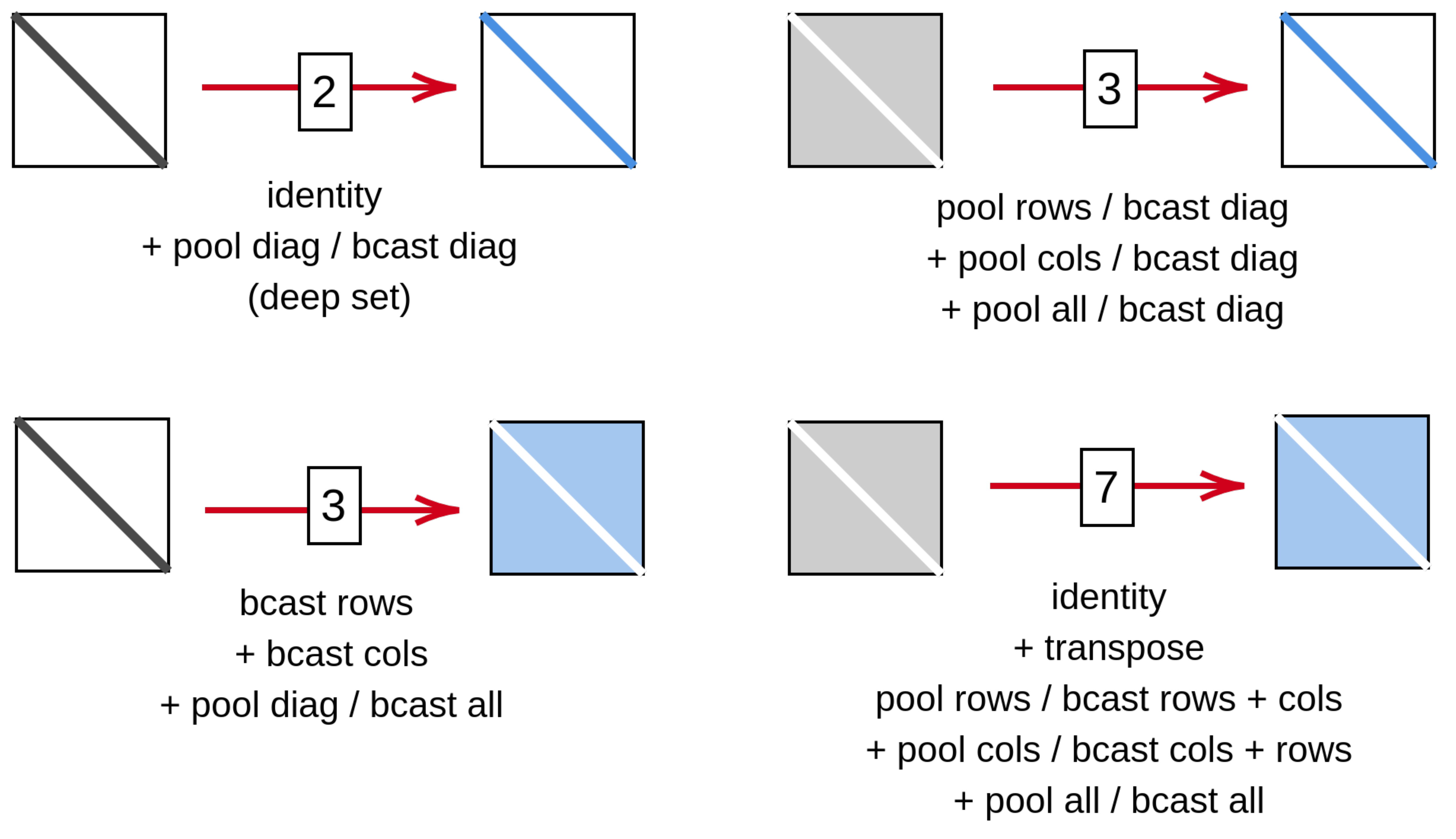}
\caption{\footnotesize{The 15 parameter equivariant map for node-node incidence matrix is decomposed into four maps between two orbits, diagonals and off-diagonals, where each orbit represents features of faces of a specific size. Each of these four maps performs pooling and broadcasting operations listed here: for example, the 3 parameter map from diagonals to off-diagonals is a linear combination of: (1) broadcasting the diagonal over rows; (2) broadcasting over columns, and; (3) first pooling the diagonal to get a scalar and broadcasting it over the off-diagonals. This approach generalizes: any $\mathcal{S}_N$-equivariant linear map between face-vectors of any size can be written as a weighted sum of all ``viable'' pool-broadcast combinations.
}
}
\label{fig:node-node-decomp}
\end{figure}

Rather than using explicit parameter-sharing in $\W$, 
we show that $\W$ is a linear combination of pooling-and-broadcasting operations. In particular, the combination includes pooling the node indices of the input face-vector in ``all possible ways'', and broadcasting the resulting collection of pooled vectors to the target face-vector, again in all ways possible. Each operation is associated with a learnable parameter; see \cref{fig:node-node-decomp} for the graph example.


\subsection{Sparse Tensors and Non-Linear Layers}
\label{sec:non-linear}
So far we have discussed equivariant linear layers for a \emph{fully connected} graph. This means dense input/output node-node incidence $\X$, or equivalently a node-edge incidence $\Y$ with the sparsity pattern described in the previous section (which is maintained by the action of $\gr{S}_{N}$). To avoid the cost of a dense representation, one may apply a sparsity mask after the linear map, while preserving equivariance:
\begin{equation}
\label{eq:non-linear}
\W_{\text{sp}}: \X \mapsto \W(\X) \circ s(\X),
\end{equation}
where $\W$ is the equivariant linear map of \cref{sec:graph-linear-layers}, $s(\X)$ is the sparsity mask, and $\circ$ is the Hadamard product. For example, assuming the layer output has the same shape as the input, we can choose to preserve the sparsity of the input. In this case, $s(\X)$ will have zero entries where the input $\X$ has zero entries, and ones otherwise.
However, the setting of \cref{eq:non-linear} is more general as input and output may have different shapes. Since the sparsity mask $s(\X)$ depends on the input, the map of \cref{eq:non-linear} is now non-linear.  In practice, rather than calculating the dense output and applying the sparsity mask, we directly produce and store only the non-zero values.

\subsection{Further Relaxation of the Symmetry Group}\label{sec:Ilayer}
The neural layers discussed so far are equivariant to the group $\gr{G}=\gr{S}_N$, where $N$ is the number of nodes. A simplifying alternative is to assume independent permutations of rows and columns of the node-node or node-edge matrix. This is particularly relevant for the node-edge matrix, where one can consider node and edges as two interacting sets of distinct objects. The corresponding $(\gr{S}_N\times \gr{S}_{N_2})$-equivariant layer, where $N_2$ is the number of edges, was introduced in \cite{hartford2018deep}, has 4 unique parameters, and it is substantially easier to implement compared to the layers introduced so far. In Appendix A we show how to construct a  sparsity-preserving (and therefore non-linear) layer for this case. Even though a single layer is over-constrained by these symmetry assumptions, we prove that two such layers generate exactly the same node and edge features as a single $\gr{S}_N$-equivariant linear layer for a node-node incidence.

\section{Higher Order Geometric Structures}
\label{sec:higher-order-objects}
In this section we generalize the results of \cref{sec:graphs} to geometric structures beyond graphs.
In \cref{sec:representation} we provide a definition of incidence tensors, which generalize node-node and node-edge matrices of graphs. We discuss several examples, showing how they can represent geometric structures such as polytopes and simplicial complexes. In \cref{sec:decomposition} we generalize the orbit decomposition of \cref{sec:graph-linear-layers}. Finally, in \cref{sec:layer}, we show how to build equivariant layers using a linear combination of simple pooling-and-broadcasting operations for arbitrary incidence tensors.

\subsection{Incidence Tensors}\label{sec:representation}
Recall that $[N] = \{1,\ldots,N\}$ denotes a set of nodes. 
 A \emph{directed face} of size $M$ is an ordered tuple of $M$ distinct nodes
$\deltas \in [N]^M \st \delta_i \neq \delta_{j} \forall i \neq j$. Following a similar logic,
an undirected face $\deltas \subseteq [N]$, is a subset of $[N]$ of size $M$.
We use $\deltas^{(M)}$ when identifying the size of the face -- \ie $|\deltas^{(M)}| = M$. For example, $\deltas^{(2)}$ identifies an edge in a graph, or a mesh, while $\deltas^{(3)}$ is a triangle in a triangulated mesh. 

An \emph{incidence tensor} $\X_{\deltas_1,\ldots,\deltas_D}$ is a tensor of order $D$, where each dimension is indexed by all faces of size $M_d = |\deltas_d|$. For example, if $\deltas_1 = \{\delta_{1,1}\}$ indexes nodes, and $\deltas_2 = \{\delta_{2,1}, \delta_{2,2}\}$ identifies an edge, $\X_{\deltas_1,\deltas_2}$ becomes a node-edge incidence matrix.  An incidence tensor has a sparsity structure, identified by a set of constraints
$\Sigma = \{\sigmas_1,\ldots,\sigmas_C\}$, where  all the indices $\sigma_{m_1}\ldots,\sigma_{m_c} \in \sigmas_c,\; \forall \sigmas_c \in \Sigma$ are equal for any non-zero entry of $\X$.
For example, we have $\X_{\{\delta_{1,1}\}, \{\delta_{2,1}, \delta_{2,2}\}} \neq 0$,
only if $\delta_{1,1} = \delta_{2,1}$. Therefore $\Sigma = \{ \sigmas_1 = \{\delta_{1,1}, \delta_{2,1}\}\}$.

While in general the pair $(\X_{\deltas_1,\ldots,\deltas_D}, \Sigma)$ defines the incidence tensor, whenever it is clear from context, we will only use $\X_{\deltas_1,\ldots,\deltas_D}$ to denote it. This formalism can represent a variety of different geometric structure as demonstrated in the following sections. 

\subsubsection{Simplicial Complexes}
Before discussing general simplicial complexes let us review graphs as an example of incidence tensors.

The \emph{node-node} incidence matrix is an incidence tensor $\X_{\{\delta_1\}, \{\delta_2\}}$ indexed by a pair of nodes, with no sparsity constraints. We denoted it with $\X_{\delta_1,\delta_2}$ in \cref{sec:graph-linear-layers} for simplicity. The \emph{node-edge} incidence matrix is denoted by the pair $(\X_{\{\delta_1\}, \{\delta_2,\delta_3\}}, \{\delta_1, \delta_2\})$. It is indexed by nodes $\{\delta_1\}$ and edges $\{\delta_2,\delta_3\}$. The entries can be non-zero only when $\delta_1 = \delta_2$, meaning that the edge $\{\delta_1, \delta_3\}$ is adjacent to the node $\{\delta_1\}$. An alternative notation is $\X_{\{\delta_1\}, \{\delta_1, \delta_2 \}}$. Again, we denoted it simply with $\X_{\delta_1,\{\delta_1, \delta_2\}}$ in \cref{sec:graph-linear-layers}. We have also denoted \emph{node} and \emph{edge} vectors with $\X_{\{\delta_1\}}$ and $\X_{\{\delta_1, \delta_2\}}$, respectively. As a final example, $\X_{\{\delta_1, \delta_2\}, \{\delta_1, \delta_3\}}$ would denote an \emph{edge-edge} incidence matrix whose entries are non-zero wherever two edges are incident.

Let us now move to the definition of a general (undirected) simplicial complex.
An \emph{abstract simplicial complex} $\Delta \subseteq 2^{[N]}$ is a collection of faces, closed under the operation of taking subsets -- that is $(\deltas_1 \in \Delta\; \text{and} \; \deltas_2 \subset \deltas_1) \Rightarrow \deltas_2 \in \Delta$.
\emph{Dimension} of a face $\deltas$ is its size minus one.
Maximal faces are called \textit{facets} and the dimension of $\Delta$ is the dimension of its largest facet. For example, an undirected graph is a one-dimensional simplicial complex. 
Each dimension of an incidence tensor $\X_{\deltas_1, \ldots, \deltas_D}$ may be indexed by faces of specific dimension. Two undirected faces of different dimension $\deltas, \deltas' \in \Delta$ are incident if one is a subset of the other.
This type of relationship as well as alternative definitions of incidence between faces of the \emph{same} dimension can be easily accommodated in the form of equality constraints in $\Sigma$.

Although not widely used, a \emph{directed} simplicial complex can be defined similarly. The main difference is that faces are \emph{sequences} of the nodes, and $\Delta$ is closed under the operation of taking a subsequence. As one might expect, the incidence tensor for directed simplicial complexes can be built using \emph{directed faces} in our notation.

\begin{example}
A zero-dimensional simplicial complex is a \emph{set} of points that we may represent using an incidence vector. At dimension one, we get \emph{undirected graphs}, where faces of dimension one are the edges. Triangulated mesh is an example of two-dimensional simplicial complex; see figure below.
\begin{center}
    \includegraphics[width=0.95\linewidth]{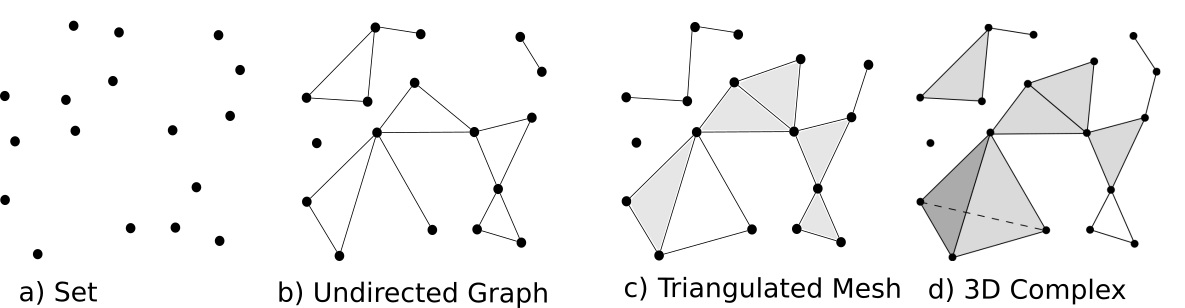}
\end{center}

The triangular bi-pyramid of \cref{fig:tetra} is an example of 3 dimensional simplicial complex with 5 nodes, 9 edges, 7 faces of size 3, and two faces of size 4.
The node-face incidence matrix in \cref{fig:tetra}(a) is expressed by $\X_{\{\delta_1\}, \{\delta_1,\delta_2,\delta_3\}}$ in our formalism. 
\end{example}

\subsubsection{Polygons, Polyhedra, and Polytopes}\label{sec:polytope}
\begin{figure}[ht]
\centering
\includegraphics[width=0.85\linewidth]{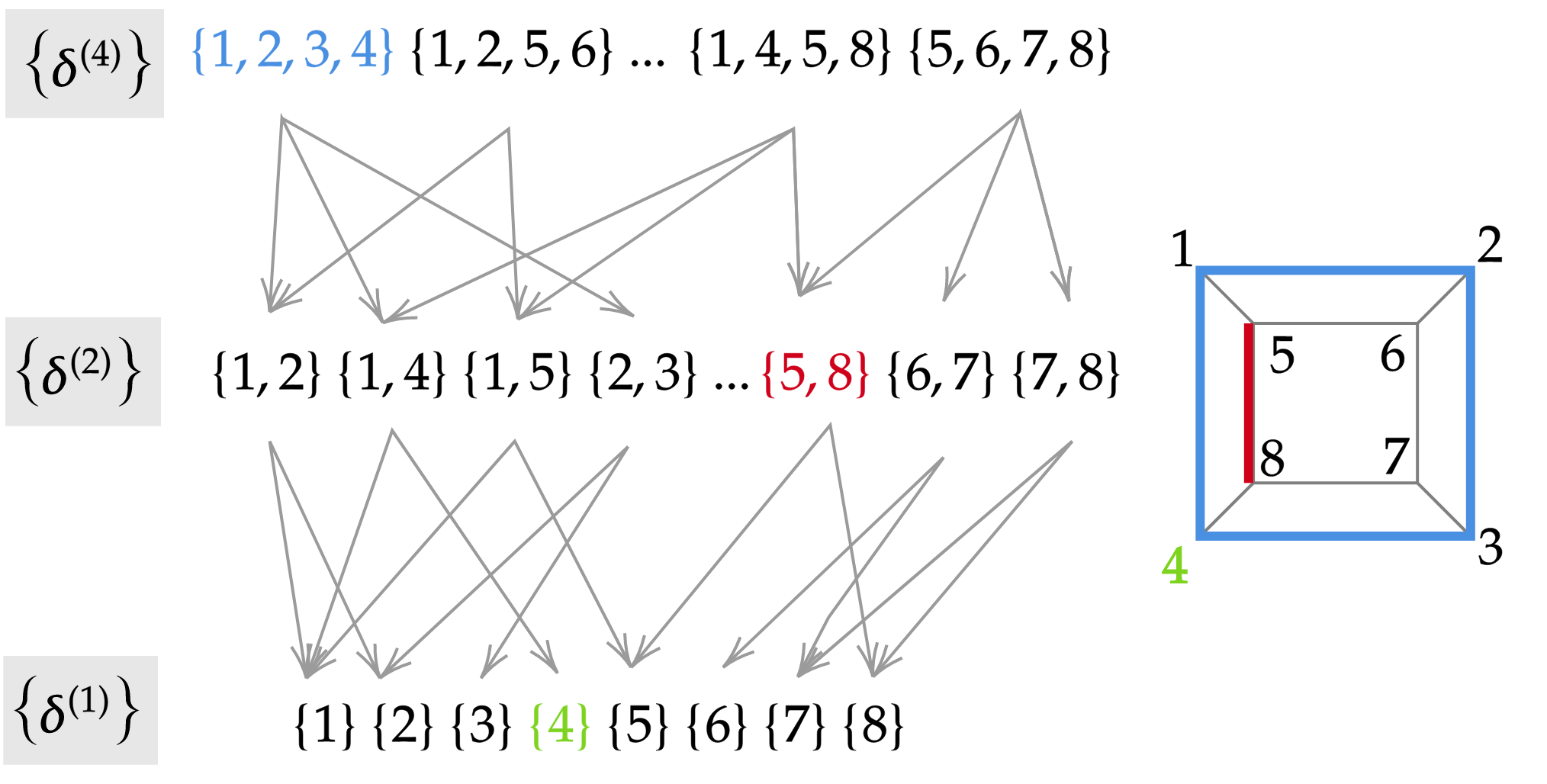}
\caption{\footnotesize{Representation of a cube as a (graded) partially ordered set. The incidence structure of the poset as well as face attributes is encoded in an incidence tensor.}}
\label{fig:poset}
\end{figure}

Another family of geometric objects with incidence structure is polytopes.
A formal definition of abstract polytope and its representation using incidence tensors is given in Appendix D. 
A \emph{polytope} is a generalization of polygone and polyhedron to higher dimensions. 
The structure of an \textit{(abstract) polytope} 
is encoded using a partially ordered set (poset) that is graded, \ie each element of the poset has a \emph{rank}. For example, \cref{fig:poset}, shows the poset for a cube, where each level is a different rank, and subsets in each level identify faces of different size (nodes, edges, and squares). 
The idea of using incidence tensor representation for a polytope, is similar to 
its use for simplicial complexes. Each dimension of $\X_{\deltas_1, \ldots, \deltas_D}$
indexes faces of different rank. Two faces of the same dimension may be considered incident if they have a face of specific lower rank in common. We may also define two faces of different dimension incident if one face is a subset of the other -- \ie $
\deltas^{(m)} < \deltas^{(m')}$ in the partial order.

\subsection{Symmetry \& Decomposition}\label{sec:decomposition}
The automorphism group $\Aut(\X) \leq \gr{S}_N$ associated with an incidence tensor is the set of all permutations of nodes that maps every face to another face, and therefore preserve the sparsity
\begin{align*}
(\X_{\pi \cdot \deltas_1, \ldots, \pi \cdot \deltas_{D}}  \neq 0  \Leftrightarrow \X_{ \deltas_1, \ldots, \deltas_{D}} \neq 0) \Leftrightarrow \pi \in \Aut(\X)   
\end{align*}
where the action of $\Aut(\X)$ on the faces is naturally defined as 
\begin{align}\label{eq:face_action}
 \pi \cdot (\delta_1, \ldots,  \delta_M) = (\pi \cdot \delta_1, \ldots, \pi \cdot \delta_M).
\end{align}
See \cref{fig:tetra}(a,b) for an example.
We may then construct $\Aut(\X)$-equivariant linear layers through parameter-sharing.
However, the constraints on this linear operator varies if our dataset has incidence tensors with different sparsity patterns. For example, a directed graph dataset may contain a fully connect graph with automorphism group $\gr{S}_{N}$ and a cyclic graph with automorphism group $\gr{C}_N$.
For these two graphs, node-node and node-edge incidence matrices are invariant to the corresponding automorphism groups, necessitating different constraints on their linear layer. 
To remedy the problem with model-sharing across instances, we \emph{densify} all incidence tensors so that all directed or undirected faces of a given dimension are present.
Now, one may use the same automorphism group $\gr{S}_N$ across all instances; see \cref{fig:tetra}(c,d). 
Next, we consider the incidence tensor as a \gr{G}-set, and identify the \emph{orbits} of $\gr{S}_N$ action.
\begin{mdframed}[style=MyFrame]
\begin{theorem}\label{th:decomposition}
The action of $\gr{S}_N$ on any incidence tensor $\X$ decomposes into orbits that are each isomorphic to a face-vector:
\begin{align}
\label{eq:decomposition}
(\X_{\deltas_1,\ldots,\deltas_D},\Sigma)
\; \cong \; \bigcupdot_{m}\; \kappa_{m}\, \X_{\deltas^{(m)}},
\end{align}
where $\kappa_m$ is the \emph{multiplicity} of faces of size $m$. The value of $\kappa_m$ is equal to the number of partitioning of the set of all indices $\{\delta_{1,1},\ldots,\delta_{1,M_1},\ldots,\delta_{D,{M_D}}\}$ into $m$ non-empty partitions, such that $\delta_{d,m}\,\forall m \in [M_d]$ belong to different partitions, and members of $\sigmas \in \Sigma$ belong to the same partition.
\end{theorem}
\end{mdframed}
The proof appears in Appendix B.
\begin{example}[Node-adjacency tensors]
\label{ex:node_adjacency}
Consider an order $D$ node-node-\ldots-node incidence tensor $\X_{\{\delta_1\},\ldots\{\delta_D\}}$ with no sparsity constraints. In this case, the multiplicity $\kappa_m$ of \cref{eq:decomposition} corresponds to the number of ways of partitioning a set of $D$ elements into $m$ non-empty subsets and it is also known as Stirling number of the second kind (written as $\stirling{D}{m}$). 
Each partition of size $m$ identifies a face-vector $\X_{\deltas^{(m)}}$ for a face of size $m$. These faces can be identified as  hyper-diagonals of order $m$ in the original adjacency tensor $\X$.
For example, as shown in the figure below, $\X_{\{\delta_1\},\{\delta_2\}, \{\delta_3\}}$ decomposes into a node-vector (the main diagonal of the adjacency cube), three edge-vectors (isomorphic to the three diagonal planes of the cube adjacency, with the main diagonal removed), and one hyper-edge-vector (isomorphic to the adjacency cube, where the main diagonal and diagonal planes have been removed). Here, $\kappa_1 = \stirling{3}{1} = 1$, $\kappa_2 = \stirling{3}{2}=3$, and $\kappa_3 = \stirling{3}{3}=1$.
\begin{center}
    \includegraphics[width=0.85\linewidth]{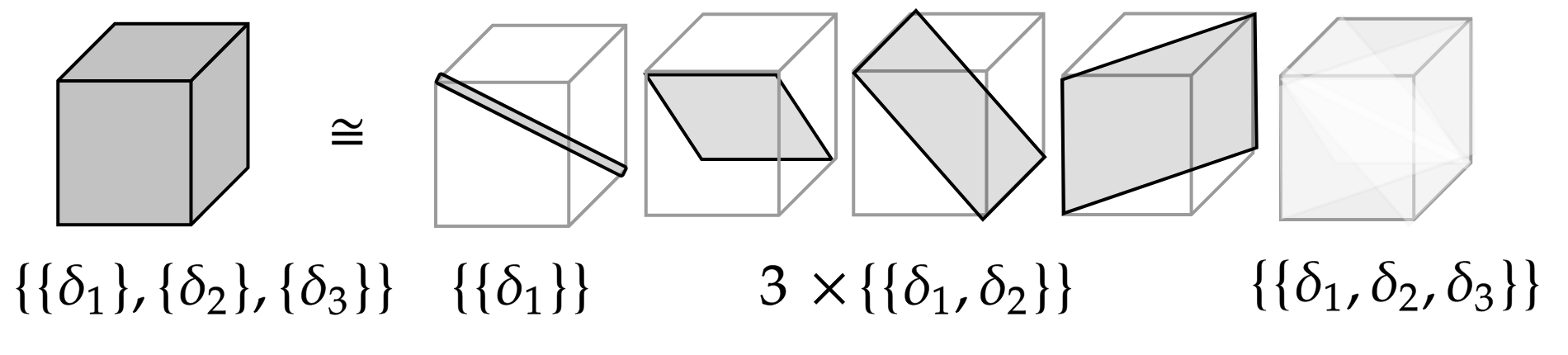}
\end{center}
\end{example}

\subsection{Equivariant Maps for Incidence Tensors}\label{sec:layer}
As shown in the previous section,
any incidence tensor can be decomposed into disjoint union of face-vectors, that are invariant sets under the action of the symmetric group.
An implication is that any equivariant map from an incidence tensor to another also decomposes into equivariant maps between face-vectors.

Let $\W^{M \to M'}$ be a linear function (here represented as a tensor) that maps a vector of faces of size $M$ to a vector of faces of size $M'$,
\begin{align}\label{eq:linear-map}
\W^{M \to M'}: \Real^{\overbracket[0.3pt]{\scriptstyle{N\times N\times\dots\times N}}^{M}}& \mapsto \Real^{\overbracket[0.3pt]{\scriptstyle{N\times N\times\dots \times N}}^{M'}}\\
 \X_{(\delta_1,\ldots,\delta_M)}&\mapsto \W^{\delta_1,\ldots,\delta_M}_{\delta'_1,\ldots,\delta'_{M'}} \X_{(\delta_1,\ldots,\delta_M)},\notag
\end{align}
where $\delta_1,\ldots,\delta_M$ identifies faces of size $M$, and (using Einstein notation) repeated indices on are summed over. Equivariance to $\gr{S}_N$ is realized through a symmetry constraint on $\W$,
\begin{equation}
\label{eq:equivariance}
\W^{\pi \cdot \delta_1,\ldots,\pi \cdot \delta_M}_{\pi \cdot \delta'_1\ldots \pi \cdot \delta'_{M'}} = \W^{\delta_1\ldots \delta_M}_{\delta'_1\ldots \delta'_{M'}} \quad \forall \pi \in \gr{S}_N,
\end{equation}
which ties the elements within each orbit of the so called \emph{diagonal} $\gr{S}_N$-action on $\W$; see \cref{fig:layer}\,(left, middle) for a graph example.

\subsubsection{Pool \& Broadcast Interpretation}\label{sec:pool_broadcast}
Each unique parameter in the constrained $\W$ corresponds to a linear operation that has a pool and broadcast interpretation -- that is any linear equivariant map 
between two incidence tensors can be written as a linear combination of pooling-and-broadcasting operations.
Moreover, this interpretation allows for a linear-time implementation of the equivariant layers, as we
avoid the explicit construction of $\W$. 
\begin{definition}[Pooling]\label{def:pool}
Given a face vector $\X_{(\delta_1\ldots \delta_M)}$, for $\set{P} = \{{p_1},\ldots,{p_L}\} \subseteq [M]$, the pooling operation sums over the indices in $\set{P}$:
\begin{align*}
 \Pool_{\set{P}} (\X_{(\delta_1\ldots \delta_M)}) = \sum_{\delta_{p_1} \in [N]} \ldots \sum_{\delta_{p_L} \in [N]}  \X_{(\delta_1,\ldots,\delta_M)},
\end{align*}
\end{definition}
In practice, the summation in the definition may be replaced with any permutation-invariant aggregation function.

\begin{definition}[Broadcasting]\label{def:broadcast} $\Bcast_{\set{B}, M'} (\X)$ broadcasts $\X$, a faces vector of size $M$, over a target vector of faces of size $M'\geq M$. We identify $\X$ with a sequence of node indices of the target face-vector, $\set{B} = (b_1,\ldots,b_M)$ with $b_m \in [M']$, and we broadcast across the remaining $M'-M$ node indices -- that is 
\begin{align*}
    \big(\Bcast_{\set{B}, M'} (\X) \big)_{(\delta_{1},\ldots, \delta_{M'})} = \X_{(\delta_{b_1},\ldots,\delta_{b_M})}.
\end{align*}
\end{definition}
For example, given an edge-vector $\X=\X_{\{\delta_1,\delta_2\}}$, $\Bcast_{\tuple{0, 1}, 3}(\X)$ broadcasts $\X$ to a triangle-vector (\ie vector of faces of size $3$), where $\X$ is mapped to the first two node indices and broadcasted along the third. 
Note that operations defined through pooling-and-broadcasting are equivariant to permutation of nodes. In fact, it turns out that an equivariant $\W$
can only linearly combine pooling and broadcasting of input incidence tensor into an output tensor. 
\begin{mdframed}[style=MyFrame]
\begin{theorem}
\label{th:pool-and-broad}
Any equivariant linear map $\W^{M \to M'}$ between face-vectors of size $M$ and $M'$,  defined in \cref{eq:linear-map}, can be written as
\begin{equation}
\label{eq:pool_broad}
\W^{M\to M'}(\X) = \sum_{\substack{\set{P}\subseteq [M]\\\set{B}\subseteq\tuple{1,\ldots,M'}\\|\set{B}| = M-|\set{P}|}} \w_{\set{B},\set{P}} \Bcast_{\set{B},M'}\big(\Pool_{\set{P}}(\X)\big).
\end{equation}
\end{theorem}
\end{mdframed}
The proof appears in Appendix B. The sum of the pooling-and-broadcasting operations in \cref{eq:pool_broad} includes pooling the node indices of the input face-vector in all possible ways, and broadcasting the resulting collection of face-vectors to the target face-vector, again in all ways possible; $\w_{\set{B},\set{P}} \in \Real$ is the parameter associated with each unique pooling-and-broadcasting combination. More details are discussed in Appendix C.

The number of operations in \cref{eq:pool_broad}, is given by
\begin{equation}
\label{eq:primitive_counting}
\tau^{M \to M'}=\sum_{m=0}^{\min(M, M')}\binom{M}{m}\binom{M'}{m}m!.
\end{equation}
This counts the number of possible choices of $m$ indices out of $M$ input indices in \cref{eq:linear-map} and $m$ indices out of $M'$ output indices to for pool and broadcast. Once this set is fixed there are $m!$ different ways to match input indices to output indices. 

\subsubsection{Decomposition of Equivariant Maps}
\label{sec:layers_generic}
Let $\W: (\bigcupdot_m \kappa_m \X_{\deltas^{(m)}}) \mapsto (\bigcupdot_{m'} \kappa_{m'}' \X_{\deltas^{(m')}})$ be an equivariant map between arbitrary incidence tensors, where both input and output decompose according to \cref{eq:decomposition}.
Using the equivariant maps $\W^{m \to m'}$ of \cref{eq:pool_broad}, we get a decomposition of $\W$ into all possible combination of input-output face vectors
\begin{align}\label{eq:general_layer}
  \W(\X_{\deltas_1,\ldots,\deltas_{D}}, \Sigma) \cong
  \bigcupdot_{m'} \bigcupdot_{k'=1}^{\kappa_{m'}'} \sum_{m} \sum_{k=1}^{\kappa_m} \W^{k,m \to k',m'} (\X_{\deltas^{(m)}}),
\end{align}
where for each copy (out of $\kappa_{m'}'$ copies) of the output face of size $m'$, we are summing over all the maps produced by different input faces having different multiplicities. Use of $k$ and $k'$ in the map $\W^{k,m \to k',m'}$ is to indicate that for each input-output copy, the map $\W^{m \to m'}$ uses a different set of parameters. 
The upshot is that input and output multiplicities $\kappa, \kappa'$ play a role similar to input and output \textit{channels}.
 
The total number of independent parameters in a layer is 
\begin{equation}
\label{eq:total_count}
    \tau = \sum_{m, m'}\kappa_{m'}'\kappa_{m} \tau^{m \to m'},
\end{equation}
where $\tau^{m \to m'}$ is given by \cref{eq:primitive_counting}.

\begin{figure*}[ht]
\centering
\includegraphics[width=.8\linewidth]{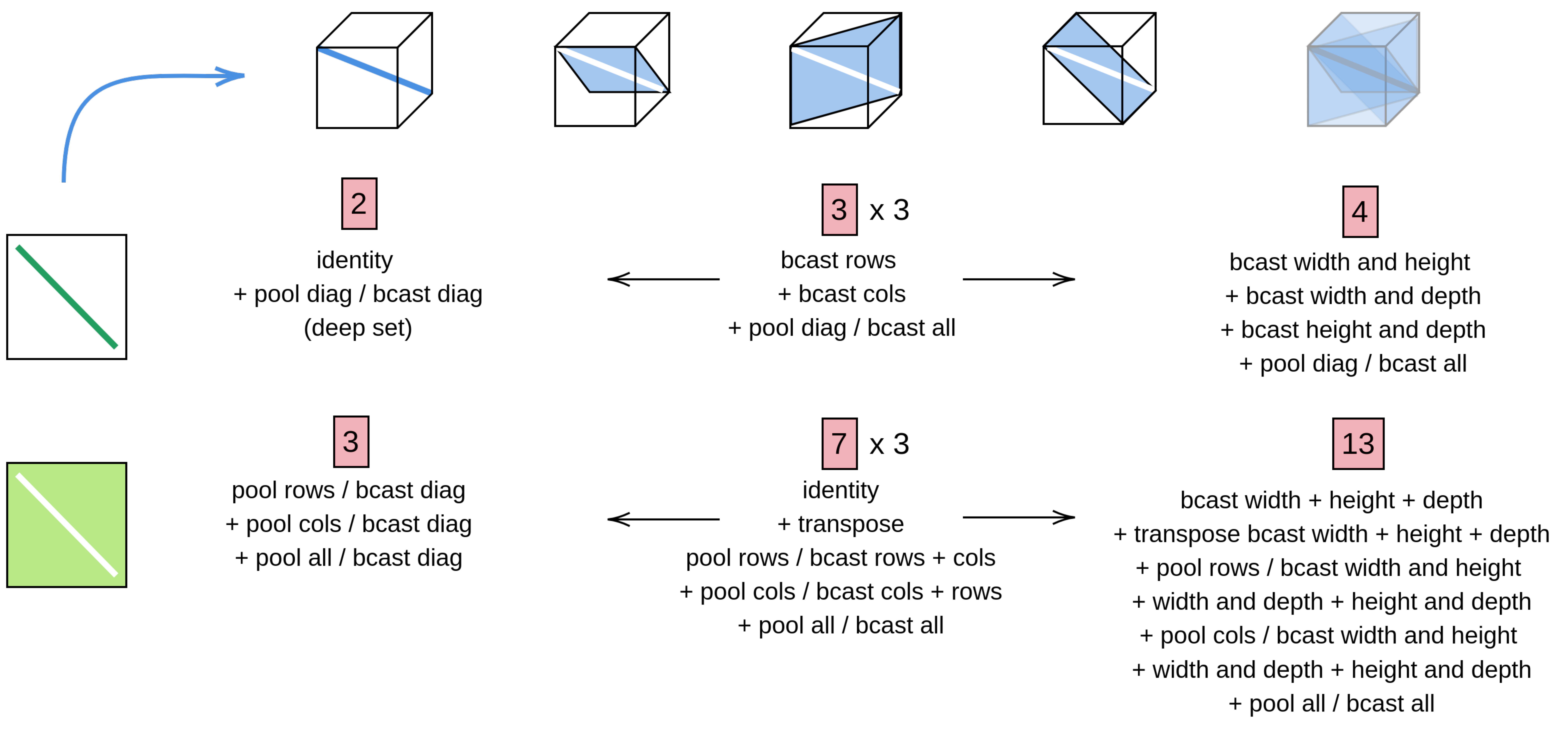}
\caption{\footnotesize{Decomposition of the $52=2 +3 + 3\times 3 + 7 \times 3 + 4 + 13$  parameter equivariant map from a node-node incidence matrix to a node-node-node incidence tensor. This figure is similar to the example of \cref{fig:node-node-decomp}, with the only difference that the output tensor has a higher rank. The input has two orbits and the output has five orbits described in \cref{ex:node_adjacency}. In our notation, the equivariant map from each orbit (face-vector) to another is $\W^{M \to M'}$ for $M = 1,2$ and $M'=1,2,3$. Each map $\W^{M \to M'}$ consists of different ways the source orbit can be pooled and broadcasted to the target orbit. 
For example, the 4 operations from a matrix diagonal to all the off-diagonals in the cube consists of broadcasting the diagonal over the cube in three different ways, plus an operation that pools over the diagonal and broadcasts the resulting scalar over the entire cube.
The number of pool-broadcast operations $\tau^{M \to M'}$ for each map is highlighted by the pink square, and the number agrees with \cref{eq:primitive_counting}. 
}
}
\label{fig:node-node-node-decomp}
\end{figure*}

\begin{example}[Node-adjacency tensors]
\label{ex:bell}
This example, is concerned with the incidence representation used in equivariant graph networks of \cite{maron2018invariant} and derives their model as a special case, using our pooling-and-broadcasting layer and face-vectors decomposition.
For an equivariant layer that maps a node-node-\ldots-node incidence tensor $\X$ of order $D$ to the same structure, the decomposition in terms of face-vectors reads
\begin{equation*}
\X_{\deltas_1,\ldots,\deltas_D} \cong \bigcupdot_{m}\stirling{D}{m}\X_{\deltas^{(m)}},
\end{equation*}
where $\stirling{D}{m}$ is the Stirling number of the second kind; see \cref{ex:node_adjacency}. The total number of operations according to \cref{eq:total_count} is then given by
\begin{align*}
 \tau &= \sum_{m, m'=1}^{D} \stirling{D}{m}\stirling{D}{m'}\sum_{l=0}^{\min(m, m')}\binom{m}{l}\binom{m'}{l}l!\\
 & = \sum_{l=0}^D\sum_{m=l}^{D}\sum_{m'=l}^{D}\bigg[\binom{m}{l}\stirling{D}{m}\bigg]\bigg[\binom{m'}{l}\stirling{D}{m'}\bigg]l! = \mathsf{Bell}(2D).
\end{align*}
In the last line, $\mathsf{Bell}(2D)$ is the Bell number and counts the number of unique partitions
of a set of size $2D$. To see the logic in the final equality: first divide $[2D]$ in half. Next, partition each half into subsets
of different sizes ($0 \leq m, m' \leq D$) and choose $l$ of these partitions from each half and merge them
in pairs. The first two terms count the number of ways we can partition each half into $m$ (or $m'$)
partitions and select a subset of size $l$ among them. The $l!$ term accounts for different ways in which $l$ partitions can be aligned. 
This result agrees with the result of \cite{maron2018invariant}.
Therefore one may implement the hyper-graph networks using efficient pooling-and-broadcasting operations outlined in \cref{eq:pool_broad}.
\end{example}

Recall that when discussing equivariant layers for graphs, we also considered independent permutations of rows and columns in a node-edge incidence matrix, and claimed that despite having only 4 parameters, stacking two such layers (with additional channels) is equivalent to the 15 parameter model. In Appendix E a similar result is given for higher dimensions, showing that one may use 
$\gr{S}_{N_{|\deltas_1|}} \times \ldots\times \gr{S}_{N_{|\deltas_D|}}$ as the symmetry group of an incidence tensor, where the equivariant model has $2^D$ parameters. 

\section{Conclusion and Future Work}
This paper introduces a general approach to learning equivariant models for 
a large family of structured data through their incidence tensor representation. In particular, we showed various incidence tensor representations for graphs, simplicial complexes, and abstract polytopes. 


The proposed family of incidence networks are 1) modular: they decompose to simple building blocks; 2) efficient: they all have linear-time pooling-and-broadcasting implementation.

In our systematic study of this family, we discussed implications of 1) added symmetry due to undirected faces; 2) sparsity preserving equivariant maps, and; 3) the  successive relaxation of the symmetry group $\Aut(\X)\leq \gr{S}_N \leq \gr{S}_{N_{|\deltas_1|}} \times \ldots\times \gr{S}_{N_{|\deltas_D|}}$, from the automorphism group, to a direct product of symmetric groups that independently permutes each dimension of the incidence tensor. Here, moving to a larger group simplifies the neural layer by reducing the number of unique parameters (and linear operations), while increasing its bias. 

Application of incidence networks to different domains, such as learning on triangulated mesh, is a natural extension of this work. Note that for a comprehensive study with nodes, edges, and triangles, as they appear in triangulated mesh, one could investigate multiple order three representations (node-node-node, node-edge-face, etc), as well as some order two representations (node-face, face-face, etc.), or simply work with face-vectors. For each of these, one potentially needs a different incidence network using a different set of sparse pool-broadcast operations. We hope to explore this direction in a follow-up work.

\section*{Acknowledgements}
We thank the reviewers for helpful comments on the manuscript. 
The computational resources for this work were mainly provided by Compute Canada. 
This research was in part supported by Canada CIFAR AI Chair Program and NSERC Discovery Grant.

\bibliography{bibliography}
\bibliographystyle{icml2020}
\clearpage
\ifbool{APX}{\appendix

\section{Additional Results for Equivariant Layers Using Relaxed Symmetry Group} \label{app:node-edge}
Consider an undirected graph with $N$ nodes and its node-node $\X_{\delta_1,\delta_2}$ and node-edge $\Y_{\delta_1,\{\delta_1,\delta_2\}}$ incidence representations. We discussed the equivalence of their $\gr{S}_N$-equivariant linear layers in \cref{sec:graph-linear-layers}. Here, we study node-edge layers equivariant to $\gr{S}_N\times\gr{S}_{N_2}$, where $N_2=N(N-1)/2$ is the number of edges, and compare their expressive power with their $\gr{S}_N$-equivariant linear counterparts. 

\paragraph{\textsc{Node-Node Incidence.}}
Let $\A\in\Real^{N\times N\times K}$ be a \emph{node-node} incidence matrix with $K$ channels. Consider a linear layer $\W_X: \A  \mapsto \A'\in\Real^{N\times N\times K'}$,
where $\W_X$ is defined as in \cref{eq:general_layer}, and $K'$ are output channels. Diagonal elements of $\A$ encode input node features and off-diagonal elements input edge features. Since the graph is undirected, $\A$ is symmetric, and the corresponding number of independent pooling and broadcasting operations (see, \eg equation \cref{eq:counting_symm})
 is $\tau_{X}=9$, for a total of $9KK'$ parameters. 

\paragraph{\textsc{Node-Edge Incidence.}}
Alternatively, one could represent the graph with a \emph{node-edge} incidence matrix $\B\in\Real^{N\times N_2\times 2K}$. Node features are mapped on the first $K$ channels along the node dimension and broadcasted across the edge dimension. Similarly, edge features are mapped on the last $K$ channels along the edge dimension and broadcasted across the node dimension. Consider a layer equivariant to $\gr{S}_N\times\gr{S}_{N_2}$ as described in 
\cref{sec:Ilayer}, that also preserves the sparsity through the non-linear implementation of \cref{sec:non-linear}, 
 $\W_Y: \B \mapsto \B'\in\Real^{N\times N_2\times 2K'}$, where $\B'=\overline{\W}_Y(\B)\circ s(\B)$, and $\overline{\W}$ is an equivariant map defined as in \cite{hartford2018deep}. We can write this linear map in our notation as
\begin{equation}
\label{eq:Ilayer}
    \overline{\W}_Y(\B) = \sum_{\set{P}\in 2^{\{0,1\}}}\lambda_{\set{P}}\Bcast_{\{0,1\}-\set{P}}\left(\Pool_{\set{P}}\left(\B\right)\right),
\end{equation}
where $2^{\{0,1\}}$ is the set of all subsets of $\{0,1\}$, and we have dropped the output dimension in the $\Bcast$ operator, as all features are broadcasted back to $\B'$. $\overline{\W}_Y(\B)$ corresponds to pooling-and-broadcasting each of the two dimensions of $\B$ independently, thus the summation has four terms, for pooling/broadcasting over rows, columns, both rows and columns and no pooling at all. The number of independent operations is $\tau_Y=4$ for a total of $16KK'$ independent parameters.

\begin{mdframed}[style=MyFrame]
\begin{theorem}
Let $\W_X$ and $\W_Y$ be the linear $\gr{S}_N$-equivariant 
and the non-linear ($\gr{S}_N\times\gr{S}_{N_2}$)-equivariant layers operating on node-node and node-edge incidence, respectively. The following statements hold:
\begin{itemize}
\item[(a)] a single $\W_Y$ layer spans a subspace of features spanned by $\W_X$,
\item[(b)] two $\W_Y$ layers span the same feature space spanned by $\W_X$ (maximal feature space).
\end{itemize}
\end{theorem}
\end{mdframed}

\begin{proof}
First, we discuss how to interpret output features. Additionally, for the rest of the proof we will assume $K=K'=1$ for simplicity, noting that the proof generalizes to the multi-channel case.
\paragraph{Output Features.} For a node-node incidence layer, it is natural to interpret diagonal and off-diagonal elements of $\A'$ as output node and edge features, respectively.

For the node-edge incidence case, all four operations of the $\W_Y$ map return linear combinations of features that vary at most across one dimension, and are repeated across the remaining dimensions. This is the same pattern of input node and edge features, and we will use the same scheme to interpret them. In particular, 
\begin{itemize}
    \item an output feature that varies across the node dimension (but it is repeated across the edge dimension) is a node feature,
    \item an output feature that varies across the edge dimension (but it is repeated across the node dimension) is an edge feature,
    \item and finally a feature that is repeated across both node and edge dimension is either a node or edge feature.
\end{itemize}
For example, consider a complete graph with three nodes. Its incidence matrix with node and edge features $\nf_{\delta_1}$ and $\ef_{\delta_1 \delta_2}$ repeat across rows and columns as follows
\begin{equation}
\begin{split}
&\B =\, \bordermatrix{
    &\sm{\myset{12}}    &\sm{\myset{13}}    &\sm{\myset{23}}\cr
\sm{1} &(\encircled{\nf_1}, \boxed{\ef_{12}})	&(\nf_1, \boxed{\ef_{13}})	    &-\\[0.3cm]\cr
\sm{2} &(\encircled{\nf_2}, \ef_{12})   & -		            &(\nf_2, \boxed{\ef_{23}})\\[0.3cm]\cr
\sm{3} &-		        &(\encircled{\nf_3}, \ef_{13})	    &(\nf_3, \ef_{23})
}\\[0.2cm]
&\parbox{10em}{$\boxed{\phantom{e}}$ = edge features\\[0.2cm]
$\encircled{\phantom{n}}$ = node features,}
\end{split}
\end{equation}
Consider pooling and broadcasting across the edge dimension. Node features were broadcasted across it, and since every node is incident with two edges, we get back multiples of the original features. The edge channel will instead return new features that combine the edges incident on each node. These new node features vary across the node dimension and are broadcasted across the edge dimension, 
\begin{equation}
\label{eq:ex_pooledges}
\resizebox{0.4\textwidth}{!}{$
\bordermatrix{
    &\sm{\myset{12}}    &\sm{\myset{13}}    &\sm{\myset{23}}\cr
\sm{1}& (2\nf_1, \ellipse{\ef_{12}+\ef_{13}})	&(2\nf_1, \ef_{12}+\ef_{13})	    &-\\[0.3cm]\cr
\sm{2}& (2\nf_2, \ellipse{\ef_{12}+\ef_{23}}) & -		                    &(2\nf_2, \ef_{12}+\ef_{23})\\[0.3cm]\cr
\sm{3} &-		                &(2\nf_3, \ellipse{\ef_{13}+\ef_{23}})	    &(2\nf_3, \ef_{13}+\ef_{23})
}.
$}
\end{equation}
On the other hand, pooling and broadcasting across the node dimension returns multiples of input edge features, and generates new edge features from the two nodes incident on each edge,
\begin{equation}
\label{eq:ex_poolnodes}
\resizebox{0.4\textwidth}{!} 
{
 \bordermatrix{
    &\sm{\myset{12}}    &\sm{\myset{13}}    &\sm{\myset{23}}\cr
\sm{1}& (\boxed{\nf_1+n_2}, 2\ef_{12})	&(\boxed{\nf_1+\nf_3}, 2\ef_{13})	    &-\\[0.3cm]\cr
\sm{2}& (\nf_1+\nf_2, 2\ef_{12})    & -		                    &(\boxed{\nf_2+\nf_3}, 2\ef_{23)}\\[0.3cm]\cr
\sm{3} &-		                &(\nf_1+\nf_3, 2\ef_{13})	    &(\nf_2+\nf_3, 2\ef_{23})
}.
}
\end{equation}
\paragraph{Proof of statement {\it (a)}.} 
Given the feature encoding described above, $\W_X$ and $\W_Y$ layers can be represented as a function acting on the space of node and edge features,
\begin{equation}
\Ga: \Real^{Q}\mapsto \Real^{Q},
\end{equation}
where $Q=N(N+1)/2$ is the number of graph elements, \ie the sum of nodes and edges. Let us fix a basis in $\Real^{Q}$ such that the first $N$ components of a vector $\phi \in\Real^{Q}$ represent node features and the remaining $N_2=N(N-1)/2$ represent edge features
\begin{equation}
\phi=\big(\underbrace{\vphantom{ \nf_{N_2}} \nf_1,\ldots, \nf_N}_{\text{node features}},\underbrace{\ef_1,\ldots, \ef_{N_2}}_{\text{edge features}}\big)^{\mathsf{T}}\equiv\big(\nfb,\efb\big)^{\mathsf{T}}.
\end{equation}
Then a layer has a matrix representation $\Ga\in\Real^{Q\times Q}$,
\begin{equation}
\label{eq:matrix-layer}
\begin{split}
\phi \mapsto \Ga \phi&=
\begin{pmatrix}
\Gb{1}{1}  &\Gb{2}{1}\\[0.2cm]
\Gb{1}{2}  &\Gb{2}{2}
\end{pmatrix}
\begin{pmatrix}
\nfb\\[0.2cm]
\efb
\end{pmatrix},
\end{split}
\end{equation}
where we have split the matrix into four sub-blocks acting on the sub-vectors of node and edge features. Here $\Gb{m}{m'}$ labels operators that maps faces of size $m$ into faces of size $m'$, with $1\leq m, m'\leq 2$. Nodes are faces of size 1 and edges faces of size 2; see also \cref{sec:higher-order-objects} for a general definition of faces. In particular, $\Gb{1}{1}\in\Real^{N\times N}$ maps input node features to output node features, $\Gb{2}{1}\in\Real^{N\times N_2}$ maps input edge features to output node features, $\Gb{1}{2}\in\Real^{N_2\times N}$ maps input node features to output edge features, and $\Gb{2}{2}\in\Real^{N_2\times N_2}$ maps input edge features to output edge features.

The nine operations of the $\W_X$ map can be written as
\begin{equation}
\label{eq:Hlayer}
\W_X \simeq
\begin{pmatrix}
\Gc{1}{1}{0}+\Gc{1}{1}{1} &\Gc{2}{1}{0}+\Gc{2}{1}{1}\\[0.2cm]
\Gc{1}{2}{0}+\Gc{1}{2}{1} &\Gc{2}{2}{0}+\Gc{2}{2}{1}+\Gc{2}{2}{2}
\end{pmatrix},
\end{equation}
and are summarized in \cref{tab:operators}. We have split the operations according to the sub-blocks defined above, where $\Gc{m}{m'}{i}$ represents operators mapping faces of size $m$ to faces of size $m'$, and with $0\leq i\leq m,m'$ representing the size of the faces after pooling. In terms of pooling-and-broadcasting of \cref{def:pool} and \cref{def:broadcast},
\begin{align}
\label{eq:Ldefinition}
\resizebox{0.48\textwidth}{!} 
{$
     \Gc{m}{m'}{i}\X_{\deltas^{(m)}} = \sum_{\substack{\set{B} \in 2^{[m']}\\ |\set{B}|=i}} \Bcast_{\set{B},m'}\left( \Pool_{[m-i]}\left( \X_{(\delta_1,\ldots,\delta_m)}\right)\right),
$}
\end{align}
where $\X_{\deltas^{(m)}}=\X_{(\delta_1,\ldots,\delta_m)}$ is a face-vector representing either node or edge features. For example, $\Gc{1}{1}{0}$ is the operator that pools all node features (\ie it pools the node-vector to dimension zero) and broadcasts the pooled scalar over nodes. The symbol $\simeq$ in \cref{eq:Hlayer} indicates that each operator is defined up to a multiplicative constant (\ie the corresponding learnable parameter $\w_{\set{B}, \set{P}}$ in \cref{eq:pool_broad}). 
The action of $\W_X$ on the space of node and edge features can be uniquely identified by the nine-dimensional subspace
\begin{equation}
\mathbb{V}_{\W_X}=\text{span}
(\left\{
\begin{pmatrix}
\Gc{1}{1}{0}   &0\\[0.2cm]
0           &0
\end{pmatrix},
\ldots,
\begin{pmatrix}
0           &0\\[0.2cm]
0           &\Gc{2}{2}{2}
\end{pmatrix}
\right\}).
\end{equation}

On the other hand, the four operations on the two channels of the $\W_Y$ map can be written as 
\begin{equation}
\label{eq:Li}
\begin{split}
\W_B &\simeq
\underbrace{
\begin{pmatrix}
\Gc{1}{1}{1} &0\\[0.2cm]
0 & \Gc{2}{2}{2}
\end{pmatrix}}_{\text{identity}}
+
\underbrace{
\begin{pmatrix}
\Gc{1}{1}{1} &\Gc{2}{1}{1}\\[0.2cm]
0 & 0
\end{pmatrix}}_{\text{pool/broadcast edges}}\\[0.2cm]
&+
\underbrace{
\begin{pmatrix}
0 &0\\[0.2cm]
\Gc{1}{2}{1} & \Gc{2}{2}{2}
\end{pmatrix}}_{\text{pool/broadcast nodes}}
+
\underbrace{
\begin{pmatrix}
\Gc{1}{1}{0} &\Gc{2}{1}{0}\\[0.2cm]
\Gc{1}{2}{0} &\Gc{2}{2}{0}
\end{pmatrix}}_{\text{pool/broadcast all}}.
\end{split}
\end{equation}
In particular,
\begin{itemize}
    \item identity: if no pooling is applied, we simply map input node and edge features into output node and edge features, respectively. Note that, since we map two input channels into two output channels, we have four independent parameters associated with this operation, but two of them are redundant. 
    \item pool/broadcast edges: when pooling and broadcasting over the edge dimension we are mapping input edge features into output node features, like the example in \cref{eq:ex_pooledges}. Input node features are also mapped into (multiples of) themselves. Similarly to the previous case, two of the four parameters are redundant. 
    \item pool/broadcast nodes: when pooling and broadcasting over the node dimension we are mapping input node features into output edge features, like the example in \cref{eq:ex_poolnodes}. Input edge features are also mapped into (multiples of) themselves. Similarly to the previous case, two of the four parameters are redundant.
    \item pool/broadcast all: when pooling and broadcasting over both dimensions we get the pooled node and pooled edge features across the entire matrix, which we can interpret as either node or edge features as described in the previous paragraph. We can use the four independent parameters associated with this operation to identify it with the operators that map pooled nodes to node and edges, and pooled edges to node and edges.
\end{itemize}

From \cref{eq:Li}, a single node-edge $\W_Y$ layer does not generate $\Gc{2}{2}{1}$, which corresponds to pooling one dimension of the edge-tensor and rebroadcasting the result to the full tensor (in order to preserve symmetry, the pooled one-dimensional tensor has to be rebroadcasted both across rows and columns). Thus, the subspace $\mathbb{V}_{\W_B}\subseteq\Real^{Q\times Q}$ of node and edge features spanned by a single $\W_Y$ map is a subspace of $\mathbb{V}_{\W_X}$,
\begin{equation}
\mathbb{V}_{\W_Y} = \mathbb{V}_{\W_X}/
\text{span}(\left\{
\begin{pmatrix}
0   &0\\[0.2cm]
0   &\Gc{2}{2}{1}
\end{pmatrix}
\right\})\subset \mathbb{V}_{\W_X}.
\end{equation}
\begin{table}[h!]
\caption{\footnotesize{The nine operators of the $\gr{S}_N$-equivariant map $\W_X$ for an undirected graph. An operator $\Gc{m}{m'}{i}$ maps faces of size $m$ to faces of size $m'$ (with $m, m' = 1$ for nodes and $m, m' = 2$ for edges), and with $i$ representing the size of partially pooled faces. Rows represent the pooled features, and columns targets each row is being broadcasted to. {\sc Partially-Pooled Edges} corresponds to pooling the $\A$ (ignoring its diagonal) either across rows or columns. {\sc Pooled Nodes} and {\sc Pooled Edges} corresponds to pooling over all node and edge features, respectively.}}
\begin{center}
\scalebox{.9}{
\begin{small}
\begin{sc}
\begin{tabular}{lcc}
\toprule
\textbf{Pooled Features / Broadcast to} & \textbf{Nodes} & \textbf{Edges}\\
\midrule
Edges                   & -                 & $\Gc{2}{2}{2}$\\
\midrule    
Nodes                 & $\Gc{1}{1}{1}$  & $\Gc{1}{2}{1}$\\[0.1cm]
Partially-Pooled Edges                     & $\Gc{2}{2}{1}$  & $\Gc{2}{2}{1}$\\
\midrule
Pooled Nodes          & $\Gc{1}{1}{0}$  & $\Gc{1}{2}{0}$\\[0.1cm]
Pooled Edges          & $\Gc{2}{1}{0}$  & $\Gc{2}{2}{0}$\\
\bottomrule
\end{tabular}
\end{sc}
\end{small}
}
\end{center}
\label{tab:operators}
\end{table}
\paragraph{Proof of statement {\it (b)}.} 
Consider the definition of $\Gc{m}{m'}{i}$ operators in \cref{eq:Ldefinition}. The following multiplication rule holds:
\begin{equation}
\label{eq:mult_rule}
\Gc{m'}{m''}{j}\Gc{m}{m'}{i} \simeq \sum_{k=\max(0, i+j-m')}^{\min(i,j)}\Gc{m}{m''}{k}.
\end{equation}
In order to show this, consider the following
\begin{equation}
\resizebox{0.48\textwidth}{!}{$
\begin{aligned}
    &\Gc{m'}{m''}{j}\left(\Gc{m}{m'}{i} \X_{\deltas^{(m)}} \right)
     = \sum_{\substack{\set{Q} \in 2^{[m'']}\\|\set{Q}|=j}} \Bcast_{\set{Q},m''}\left( \Pool_{[m'-j]}\left(\Gc{m}{m'}{i} \X_{\deltas^{(m)}} \right)\right) \\
    & = \sum_{\substack{\set{Q} \in 2^{[m'']}\\|\set{Q}|=j}} \Bcast_{\set{Q},m''}\left( \Pool_{[m'-j]} \left (\sum_{\substack{\set{P} \in 2^{[m']}\\|\set{P}|=i}} \Bcast_{\set{P},m'}\left( \Pool_{[m-i]}\left( \X_{\deltas^{(m)}} \right )\right)\right)\right)\\
    & \simeq \sum_{k=\max(0,i+j-m')}^{\min(i,j)} 
    \sum_{\substack{\set{R} \in 2^{[m'']}\\|\set{R}|=k}} \Bcast_{\set{R},m''}\left( \Pool_{[m''-k]}\left( \X_{\deltas^{(m)}}\right)\right)\\
    & =  \sum_{k=\max(0, i+j-m')}^{\min(i,j)}\Gc{m}{m''}{k} \X_{\deltas^{(m)}},
\end{aligned}
$}
\end{equation}
where the first two lines are simply substituting the definition of the operators. The important step is going from second to third line: this is considering all the pooled tensors that can be created and broadcasting them to the target $m''$-dimensional tensor. Note that only the dimension of these pooled tensors is important as the constants are irrelevant and absorbed in the $\simeq$ symbol. In particular, the lowest dimension $\X_{\deltas^{(m)}}$ is pooled to, is  $\max(0, m - (m-i) - (m'-j)) = \max(0, i+j-m')$.
On the other hand, the maximum dimension $\X_{\deltas^{(m)}}$ is pooled to is $\min(i, j)$. Also, note that \cref{eq:mult_rule} is consistent with the special case $m'=m''=j$. In this case $\Gc{m'}{m'}{m'}\simeq \boldsymbol{1}$ is proportional to the identity operator. Consistently, we get $\Gc{m'}{m'}{m'}\Gc{m}{m'}{i}\simeq \sum_{k=\max(0,i)}^{\min(m',i)}\Gc{m}{m'}{k}=\Gc{m}{m'}{i}$.
The same relation holds for the special case $m=m'=i$.

Using \cref{eq:mult_rule} we get
\begin{equation}
\begin{split}
\W_Y^2&\simeq
\begin{pmatrix}
\Gc{1}{1}{0}+\Gc{1}{1}{1} &\Gc{2}{1}{0}+\Gc{2}{1}{1}\\[0.2cm]
\Gc{1}{2}{0}+\Gc{1}{2}{1} &\Gc{2}{2}{0}+\Gc{2}{2}{1}+\Gc{2}{2}{2}
\end{pmatrix}\\
&\simeq \W_X \implies \mathbb{V}_{\W_Y^2} = \mathbb{V}_{\W_X}.
\end{split}
\end{equation}
Two stacked $\W_Y$ maps span the same subspace of operators and thus output features as a single $\W_X$ map. Note that $\Gc{2}{2}{1}$ missing from a single $\W_Y$ is generated by composing $\Gc{2}{2}{1}\simeq\Gc{1}{2}{1}\Gc{2}{1}{1}$, that is edge features are pooled to dimension one and broadcasted to nodes in the first layer through $\Gc{2}{1}{1}$, and then re-broadcasted across rows and columns, like all the other node features, in the second layer through $\Gc{1}{2}{1}$. Furthermore, using the same multiplication rules, we find that
\begin{equation}
\W_X \simeq \W_X^m\simeq \W_Y^{1+m}\quad\forall m\in\Nat\st m\geq 1,
\end{equation}
thus by taking a single $\W_X$ map or by stacking two $\W_Y$ maps we span the maximal node and edge feature space.
\end{proof}

\section{Proofs}
\label{app:proofs}
\subsection{Proof of \cref{th:decomposition}}
\begin{proof}
Consider an incidence tensor $(\X_{\deltas_1,\ldots,\deltas_D},\Sigma)$. Assume there exist two node indices $\delta_{d,m}$, $\delta_{d',m'}$ within two face indices $\deltas_{d}$ and $\deltas_{d'}, d' \neq d$ that are not constrained to be equal by $\Sigma$; therefore they can be either equal or different.
The action of $\gr{S}_N$ (and any of its subgroups) maintains this (lack of) equality, that is 
\begin{align*}
 \pi \cdot \delta_{d,m} = \pi \cdot \delta_{d', m'} \Leftrightarrow \delta_{d,m} = \delta_{d', m'} \forall \pi \in \gr{S}_N.   
\end{align*}

This means that the set of indices constrained by $\Sigma$ $\{\deltas_1,\ldots,\deltas_D \mid \Sigma\}$, can be partitioned into two disjoint G-sets 
\begin{align*}
    \{\deltas_1,\ldots,\deltas_D \mid \Sigma\} = &\{\deltas_1,\ldots,\deltas_D \mid \Sigma, \delta_{d,m} = \delta_{d',m'}\} \cupdot \\ &\{\deltas_1,\ldots,\deltas_D \mid \Sigma, \delta_{d,m} \neq \delta_{d',m'}\},
\end{align*}
with $\delta_{d,m} = \delta_{d',m'}$ in one G-set and $\delta_{d,m} \neq \delta_{d',m'}$ in the other. We may repeat this partitioning recursively for each of the resulting index-sets. This process terminates with \emph{homogeneous} G-sets
 where any two indices $\delta_{d,m}$ and $\delta_{d',m'}$ are either constrained to be equal or different. It follows that if we aggregate all equal indices, we are left with a set of indices that are constrained to be different, and therefore define a face $\deltas^{(m)}$.

The number of ways in which we are left with $m$ 
different indices $\kappa_m$ is given by the number of partitions of $\{\delta_{1,1},\ldots,\delta_{1,M_1},\ldots,\delta_{D,M_D}\}$ into $m$ non-empty sets, where 
two elements in the same partition are constrained to be equal. This partitioning is further constrained by the fact that elements within the same face $\deltas_d$, are constrained to be different and therefore should belong to different partitions. Moreover, $\Sigma$ adds its equality constraints.
\end{proof}

\subsection{Proof of \cref{th:pool-and-broad}}
\begin{proof}
Consider the linear map $\W^{M\to M'}$ of \cref{eq:linear-map}. Each unique parameter corresponds to an orbit under the action of $\gr{S}_N$. Orbits are partitions of the set  $\{\delta_1,\ldots,\delta_M,\delta'_1,\ldots,\delta'_{M'}\}$ of input and output node indices, where any subset of the partition contains at most one input and one output index. That is, subsets have either size one, with a single input or output index, or size two, with a pair of input and output indices. Given such a partition, all and only $\W^{M \to M'}$ elements such that indices in the same subset are equal and indices in separate subsets are different, identify an orbit. That is, given any two elements of $\W^{M \to M'}$, there exists an element of $\gr{S}_N$ that maps one onto the other iff they satisfy the equality pattern described above. Each partition can be mapped to a pooling-and-broadcasting operation as detailed in \cref{app:pool-broad}.

This is similar to the construction of \cite{maron2018invariant}. Here, we generalize their partitions approach and the pooling-and-broadcasting implementation to arbitrary incidence tensors (through the decomposition of \cref{th:decomposition}).
\end{proof}

\section{Pooling-and-broadcasting from Partitions} \label{app:pool-broad}
Orbits of an equivariant linear map $\W^{M\to M'}: \X_{(\delta_1,\ldots,\delta_M)}\mapsto \X_{(\delta'_1,\ldots,\delta'_{M'})}$ are partitions of the set  $\{\delta_1,\ldots,\delta_M,\delta'_1,\ldots,\delta'_{M'}\}$ of input and output node indices, where any subset of the partition contains at most one input and one output index, as described in \cref{app:proofs}. Note that for the partitions of interest, subsets have either size one, with a single input or output index, or size two, with a pair of input and output indices. We map each partition to the corresponding pooling-and-broadcasting operation: 
\begin{itemize}
\item[(a)] input indices in subsets of size one are pooled over,
\item[(b)] input and output indices in the same subsets are mapped onto each other,
\item[(c)] output indices in subsets of size one are broadcasted over.
\end{itemize}
\begin{example}
Consider the function $\W^{2\to 1}$ that maps faces of size two (edges) to faces of size one (nodes). We introduced this map, \eg in \cref{sec:graph-linear-layers}. From \cref{eq:primitive_counting} $\W^{2\to 1}$ has three orbits. Let $\delta_1,\delta_2$ be input node indices and $\delta'_1$ the output node index. Following the rules outlined above we get:
$\{\{\delta_1\}, \{\delta_2\}, \{\delta'_1\}\}$ corresponds to pooling both edge dimensions and broadcasting the result to the output nodes, 
$\{\{\delta_1, \delta'_1\}, \{\delta_2\}\}$ corresponds to pooling the edge second dimension and mapping the resulting one-dimensional vector to output nodes, finally $\{\{\delta_2, \delta'_1\}, \{\delta_1\}\}$ corresponds to pooling the edge first dimension and mapping the resulting one-dimensional vector to output nodes.
\end{example}

\section{More on Abstract Polytopes}\label{app:polytope}
The structure of an \textit{abstract polytope} 
is encoded using a \emph{graded} partially ordered set (poset).
A poset is a set equipped with a partial order that enables transitive comparison of certain members. A poset $\Pi$ is graded if there exists a rank function $\operatorname{rank}: \Pi \mapsto \mathbb{N}$ satisfying the following constraints:
\begin{align*}
&\deltas < \deltas' \Rightarrow \operatorname{rank}(\deltas) < \operatorname{rank}(\deltas') \quad \forall \deltas, \deltas' \in \Pi \\
&\not \exists \deltas'' \in \Pi \; s.t. \; \deltas < \deltas'' < \deltas' 
\Rightarrow
\operatorname{rank}(\deltas') = \operatorname{rank}(\deltas) + 1
\end{align*}
An abstract polytope is a set $\Pi$ of partially ordered faces of different dimension. In a geometric realisation, the partial order is naturally defined by the inclusion of a lower-dimensional face in a higher dimensional one (\eg an edge that appears in a face of a cube). 
\cref{fig:poset} shows the partial order for a cube, where we continue to use a set of nodes to identify a face.

We can define the incidence structure similar to simplicial complexes. For example, we may assume that two faces $\deltas, \deltas' \in \Pi$ of different dimension (rank)
are incident iff $\deltas < \deltas'$, or $\deltas > \deltas'$. Similarly, we may assume that two faces $\deltas,\deltas'$ of the same dimension $d$ are incident iff there is a face $\deltas''$ of dimension $d-1$ incident to both of them $\deltas'' < \deltas, \deltas'$. Note that if the polytope is \emph{irregular}, faces of similar rank may have different sizes -- \eg consider the soccer ball where pentagons and hexagons have the same rank.

\section{Additional Results for Higher Order Geometric Structures}\label{app:additional-results}
\subsection{Additional Symmetry of Undirected Faces}
When counting the number of unique pooling and broadcasting operations, we assumed that
$\X_{\deltas^{(M)}} = \X_{(\delta_1,\ldots,\delta_M)} \neq \X_{(\pi \cdot \delta_1,\ldots,\pi \cdot \delta_M)}$. However, if $\deltas^{(M)}$ is an undirected face, the face-vector is invariant to permutation of its nodes.
 This symmetry reduces the number of independent parameters in \cref{eq:pool_broad}.  Furthermore, if we enforce the same symmetry on the output face-vector, some of the weights need to be tied together. In particular, there is only one vector of faces of size $m$ that can be extracted through pooling for each value of $0 \leq m \leq \min(M,M')$. Similarly, all possible ways of broadcasting the pooled face-vector to the target face-vector have to carry the same parameter in order to restore symmetry of the output. Thus, in comparison to  \cref{eq:primitive_counting}, for the symmetric input and symmetric output case, the degrees of freedom of the equivariant map are significantly reduced:
\begin{equation}
\label{eq:primitive_counting_symm}
\tau^{(M \to M')}_{\text{symm}} = \sum_{m=0}^{\min(M,M')} 1 = \min(M, M') + 1.
\end{equation}

\begin{example}[Symmetric node-adjacency tensors]
Consider a similar setup, but for the case of undirected faces.
In this case we map a symmetric input into a symmetric output. From \cref{eq:total_count}, where we use now \cref{eq:primitive_counting_symm} for the symmetric case, we get
\begin{equation}
\label{eq:counting_symm}
\tau = \sum_{m,m'=1}^D(\min(m, m') + 1) = \frac{1}{6}(2D^3+9D^2+D).
\end{equation}
Note that we have omitted the multiplicity coefficients $\kappa_{m,m'}$, assuming that all faces of a given dimension are equal.
\end{example}

\subsection{Representation and Expressiveness}
Consider a simplicial complex or polytope where faces of particular dimension have associated attributes. This information may be directly represented using face-vectors $\bigcupdot_{m=1}^{M} \X_{\deltas^{(m)}}$. 
Alternatively, we may only use the largest face $\deltas^{(M)}$, and broadcast all lower dimensional data to the maximal face-vectors. This gives an equivalent representation 
\begin{align*}
 \bigcupdot_{m=1}^{M} \X_{\deltas^{(m)}} \equiv \bigcupdot_{m=1}^{M} \Bcast_{\set{B},M}\left( \X_{\deltas^{(m)}}\right) = \bigcupdot_{k=1}^{M} \X^k_{\deltas^{(M)}},
\end{align*}
that resembles having multiple channels, indexed by $k$.

Yet another alternative is to use an incidence tensor $\X_{\deltas_1,\ldots,\deltas_M \mid \Sigma} \cong \bigcupdot_{m=1}^{M} \kappa_m \X_{\deltas^{(m)}}$ that decomposes into face vectors according to \cref{th:decomposition}. We have similarly diverse alternatives for the ``output'' of an equivariant map. \emph{Observe that the corresponding equivariant maps have the same expressiveness up to change in the number of channels.} 
This is because we could produce the same pool-broadcast features of \cref{eq:pool_broad} across different representations. 

An example is the node-edge incidence matrix discussed in \cref{sec:graph-linear-layers}, where node features are broadcasted across all edges incident to each node and whose equivariant layer has the same expressiveness as the layer for the corresponding node-node matrix.

\subsection{Non-Linear Layers and Relaxation of the Symmetry Group}
Non-linear layers, where the sparsity of the output is controlled by an input-dependent mask as in \cref{eq:non-linear}, are easily generalized to arbitrary incidence tensors.

Similarly, as discussed in \cref{sec:Ilayer} for a node-edge incidence matrix, one can consider independent permutation for each dimension of the incidence tensor. Let $N_{m} = |\{\deltas^{(m)}\}|$ denote the number of faces of size $m$, so that $N_1 = N$. 
Consider the action of $\boldsymbol{\pi} = (\pi^1\ldots,\pi^D) \in \gr{S}_{N_{|\deltas_1|}} \times \ldots\times \gr{S}_{N_{|\deltas_D|}}$ on $\X_{\deltas_1,\ldots,\deltas_D}$:
\begin{align*}
    \boldsymbol{\pi} \cdot \X_{\deltas_1,\ldots,\deltas_D} = \X_{\pi^1 \cdot \deltas_1,\ldots,\pi^D \cdot \deltas_D},
\end{align*}
where $\pi^d \cdot \deltas_d$ is one of $N_{|\deltas_d|}!$ permutations of these faces.
The corresponding equivariant layer introduced in \cite{hartford2018deep} has $\tau = 2^{D}$ unique parameters.

}{}
\end{document}